# RetroGNN: Approximating Retrosynthesis by Graph Neural Networks for De Novo Drug Design

Cheng-Hao Liu[1,4], Maksym Korablyov[1], Stanisław Jastrzębski[2,5], Paweł Włodarczyk-Pruszyński[2], Yoshua Bengio[1], and Marwin H. S. Segler[3]

[1]Mila and Université de Montréal, Canada
[2]Molecule.one, Poland
[3]Westfälische Wilhelms-Universität Münster, Germany
[4]McGill University, Canada
[5]Jagiellonian University, Poland
*chenghao.liu@mail.mcgill.ca, mkorablyov@gmail.com, stan@molecule.com, pawel@molecule.one, yoshua.bengio@mila.quebec, marwin.segler@wwu.de*

## Abstract

De novo molecule generation often results in chemically unfeasible molecules. A natural idea to mitigate this problem is to bias the search process towards more easily synthesizable molecules using a proxy for synthetic accessibility. However, using currently available proxies still results in highly unrealistic compounds. We investigate the feasibility of training deep graph neural networks to approximate the outputs of a retrosynthesis planning software, and their use to bias the search process. We evaluate our method on a benchmark involving searching for drug-like molecules with antibiotic properties. Compared to enumerating over five million existing molecules from the ZINC database, our approach finds molecules predicted to be more likely to be antibiotics while maintaining good drug-like properties and being easily synthesizable. Importantly, our deep neural network can successfully filter out hard to synthesize molecules while achieving a $10^5$ times speed-up over using the retrosynthesis planning software.

## 1 Introduction

The number of drug-like molecules is estimated at $10^{23}$-$10^{60}$ [1]. Despite decades of high-throughput screening, only a tiny fraction of this chemical space has been surveyed. De novo molecule generation aims to efficiently explore the vast drug-like universe and generate molecules *from scratch* via computational methods [2].

Unconstrained de novo design methods often generate unrealistic and hard to synthesize molecules [3, 4]. Accordingly, discerning between easy and hard to synthesize molecules has long been an area of focus for the field. Several approaches have shown promise for improving the synthesizability of molecules proposed by de novo models. Perhaps the most natural idea is to bias the search towards easier to synthesize compounds using a synthesizability scoring function such as the SAScore or SCScore [5, 6, 7]. However, current scores trade accuracy for speed, and their use can still result in highly unrealistic compounds [4].

Another approach is to search in a space where molecules are constructed by chemical reactions [8, 9, 10, 11]. For example, [10] train a reinforcement learning agent that in each step modifies the molecule by applying one of several predefined reaction templates. This leads to feasible molecules; however, chemically very similar molecules could be produced by very different synthetic routes making optimization in this space potentially more challenging.

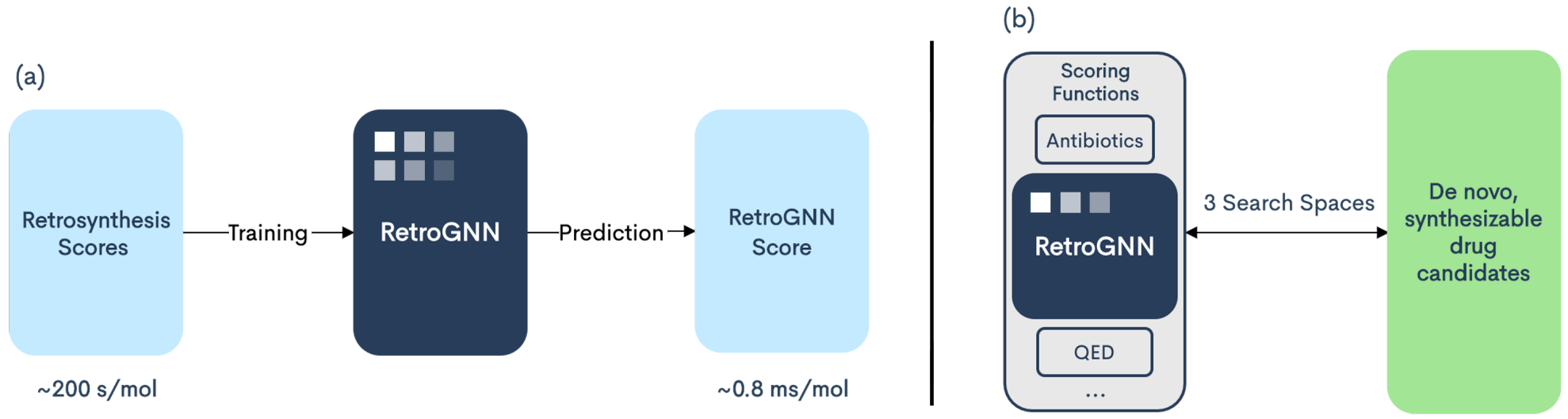

Figure 1: Workflow schematics in this paper. **(a)** We approximate a retrosynthesis planning software outputs using a graph neural network (RetroGNN). In the considered setting, we achieved approximately $10^5$ speed-up (0.8 ms / molecule on a single GPU compared to 200s / molecule achieved by Molecule.one public API on November 2020 when using a single CPU core). **(b)** To demonstrate the effectiveness of RetroGNN in-silico, we search for easily synthesizable antibiotics candidates using the RetroGNN score as one of the components of the scoring function.

To avoid constraining the search space, we propose a novel approach based on retrosynthesis planning software. Recent work has shown promising results in automatic synthesis planning using deep learning and search [12, 13, 14, 15]. However, these methods are generally too slow (typically tens of seconds to minutes per molecule [14]) to be used in the large number of evaluations required for de novo drug design [4]. To provide the necessary speed-up for scoring during de-novo design, we propose to train a deep neural network to predict the outputs of a retrosynthesis planning software.

Our workflow is the following: (1) We designed three search/action spaces to generate molecules. (2) In each search space, we trained an accurate and efficient message-passing neural network on outputs of a retrosynthesis planning software [16]. (3) We constructed a score to reflect an estimate of synthesizability, drug-likeness, and the probability of being an antibiotic. (4) We validated the approach by searching for synthesizable, drug-like molecules with antibiotic properties [17].

## 2 Synthetic Accessibility Score by approximating a Retrosynthesis Planner

**M1Score** To train a neural network to approximate outputs of a synthesis planning software, we experiment with the synthetic accessibility score provided by the Molecule.one retrosynthesis planning software (M1Score). For a given molecule, the score provides a number between 1 and 10 that reflects the expected cost of making the molecule, and 11 indicates a lack of synthesis routes.[1]

Similarly to [18, 19], the Molecule.one software combines search and deep neural networks to automatically plan synthesis routes. The search is also guided by the prices of available compounds. We provide further details on M1Score, as well as compare M1Score to SAScore, in Appendix 4.1.

**Approximating M1Score** Synthesis planning software are generally too slow to apply to de novo drug design within the optimization loop [4]. To alleviate this issue, we train a deep neural network to predict the M1Score. Specifically, we train deep message passing neural networks (MPNN) [20] to predict the M1Score from the molecular structure, to which we refer the model and the score as RetroGNN and RetroGNNScore, respectively. We experimented with different MPNN configurations from [20] and [21], and tuned their hyperparameters. RetroGNN were trained using 50k molecules sampled randomly from each search space. The details are in Appendix 4.2

## 3 Experiments

We seek to understand (1) how accurate and how fast the developed score can predict the outcome of a synthesis planner and (2) how well the score can bias the search towards molecules that are deemed to be easy to synthesize by the retrosynthesis planning software together with other objectives.

[1]Please note that the method of approximating the outcome of a synthesis planner using ML is agnostic to the specific implementation of synthesis planner used, and is readily adaptable to other planning software as well.

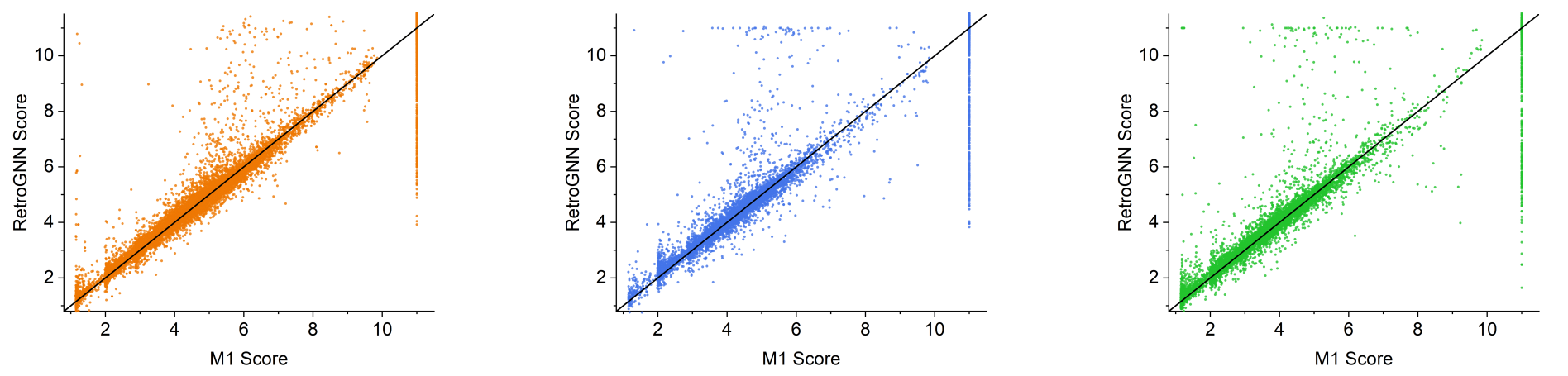

(a) 105 fragments search space (b) 464 fragments search space (c) Graph edit search space

Figure 2: RetroGNNScore approximates M1Score well on randomly generated molecules from the three considered search spaces, achieving $R^2 = 0.982, 0.976, 0.995$, respectively from (**a**-**c**).

### 3.1 Experimental setup

We experiment with three different molecular search spaces. Each search space is defined by a set of modifications that can be applied to the molecule. We use two building block search spaces that respectively consist of 105 and 464 most common fragments in the PDB database, where an action either connects or disconnects two substructures [22, 23]. We also perform the same experiment with a graph edit search space, where the actions include: adding an atom, mutating an atom/bond, deleting an atom/bond, and adding an aliphatic ring or an aromatic ring, similarly to [24].

To search, we initialize each trajectory with a random molecule, and we proceed for 200 steps to optimize this molecule. Each optimization step proceeds as follows: (1) estimating the value of each action by evaluating the score of the resulting molecules (2) taking one of the actions randomly according to the Boltzmann distribution obtained by exponentiating and normalizing the scores (softmax action selection) [25]. Experimental details are described in Appendix 4.3.

### 3.2 How well and how fast does RetroGNN approximate a synthesis planner?

Our key motivation for introducing RetroGNNScore was to make using synthesis planning feasible for de novo drug design. We found that RetroGNNScore is ~$10^5$ times faster to compute than M1Score; on a single GPU, this enables the screening of a billion molecules in 9 days compared to otherwise 6000 years. We include details on this comparison in Appendix 2.

To examine how well RetroGNNScore approximates M1Score, in each environment/action space, we study the correlation between RetroGNNScore and M1Score for 50,000 randomly sampled molecules. Figure 2, Figure 7, and Figure 8 show the outcome. For binary classification of whether there are any synthetic pathways found by Molecule.one, we achieve an AUC score of 0.995-0.998. For regression, we observe an excellent correlation ranging from 0.976 to 0.995 in $R^2$. A lower, but still satisfactory, correlation of 0.910 and 0.941 $R^2$ is observed for FDA-approved and investigational drugs.

### 3.3 Finding synthesizable and drug-like antibiotics

We evaluate the effectiveness of our method *in silico* on a multi-objective optimization task by searching for drug-like molecules with antibiotic activity. We utilize the model ($p(a_\theta \mid x)$) published by [17], which predicts the probability of compound $x$ suppressing the activity of *E. coli*. Specifically, we search for a molecule $x$ that *minimizes*:

$$\text{score}(x) = \text{AntibioticScore}(x) \cdot \text{QEDScore}(x) \cdot \text{SynthScore}(x), \tag{1}$$

where

$$\text{AntibioticScore}(x) = \log(1 - p(a_\theta \mid x)), \tag{2}$$

$$\text{QEDScore}(x) = \min(\max(0.0, \frac{\text{QED}(x)}{0.7}), 1.0), \tag{3}$$

$$\text{SynthScore}(x) = \min(\max(0.0, \frac{11 - \text{RetroGNNScore}(x)}{11 - 4.5}), 1.0), \tag{4}$$

$$\text{or, SynthScore}(x) = \min(\max(0.0, \frac{11 - \text{SAScore}(x)}{11 - 3.5}), 1.0). \tag{5}$$

Equation 1 can be seen as the optimization of $\text{AntibioticScore}(x)$ with linear constraints of $\text{QEDScore}(x)$ and $\text{SynthScore}(x)$. Here, QED is the Quantitative Estimate of Drug-likeness [26]. For synthesizablity, we either use the baseline SAScore [7] or RetroGNNScore in $\text{SynthScore}(x)$.

To obtain proper constraints, we calculated the average QED, M1Score, and SAScore in FDA-approved and investigational drugs (ZINC15 database [27]) to be 0.52, 5.0, and 3.5, respectively. We consider performance better than these to be sufficient for drug discovery, and hence we cap the optimization for QED, RetroGNNScore, and SAScore at 0.70, 4.5, and 3.5, respectively (Eq. 3,4,5).

We employ a heuristic approach where the algorithm visits each action proportionally to the normalized probability of the resulting molecule being an antibiotic. As $P \sim exp(reward)$ in Boltzmann sampling, we minimize $log(1 - p(a_\theta \mid x))$, which can be thought of as taking logits from the last layer of the classifier network, and where we obtain exponential reward as $p(a_\theta \mid x) \to 1$.

As $\text{AntibioticScore}(x)$ is strictly negative (lower is better), whereas $\text{QEDScore}(x)$, $\text{SynthScore}(x)$ are strictly positive (higher is better), minimizing Equation 1 optimizes all three properties.

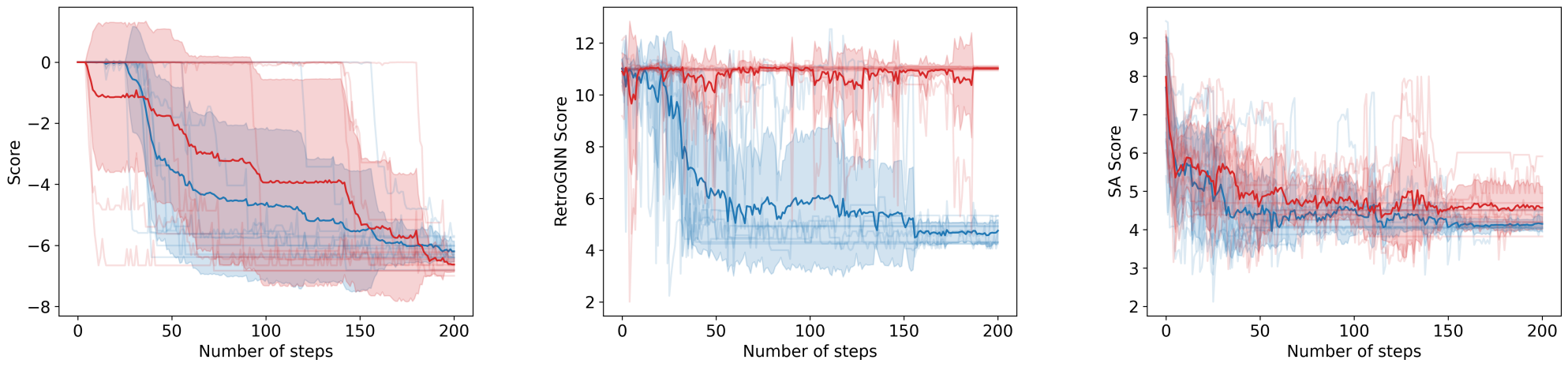

(a) Overall score (b) RetroGNNScore (c) SAScore

Figure 3: Softmax action selection with RetroGNNScore can find molecules with higher likelihoods of antibiotic activity while optimizing for synthesizability in comparison to using virtual library screening. Top 10 trajectories and their average from softmax action selection in the 464 fragments search space that optimizes for the score in Equation 1. Color indicates whether we optimize using SAScore (red) or RetroGNNScore (blue). Optimizing for RetroGNNScore also reduces SAScore.

**Results** We first show the need for effective de novo generation of synthesizable molecules by comparing against several baselines[2]. We begin by searching through 5.9 million known molecules in the drug-like subset of the Zinc15 database[27]. We find the top-1 and top-$100^{\text{th}}$ scores be -3.6 and -2.5, respectively. These molecules generally possess high QED and M1Score, but such a large dataset only produces molecules with moderate estimated probability of antibiotic activity. On the other hand, we generate 1.5 million new molecules via random actions in the three different search spaces, where the majority of the molecules (>80%) end up being unsynthesizable, and thus the scores for synthesizable yet antibiotic-active molecules are also correspondingly low. These results are shown in Figure 9.

Next, we investigate using softmax action selection to find high-scoring molecules in each considered search space. Using the optimal temperature tuned to be 0.15, we observe that employing RetroGNNScore biases the search towards easy to synthesize molecules according to M1Score, with high QED values. With just 1000 searches, our model finds molecules predicted to be more likely an antibiotic than those found by screening the ZINC database (and hence [17]) and random search by up to $-7$ in $\text{AntibioticScore}(x)$ while maintaining high synthesizability. The trajectories show the optimizer explores local unfavorable actions that eventually lead to higher overall scores.

Interestingly, such optimization naturally leads to good SAScore, but using SAScore in lieu of RetroGNNScore finds molecules predicted to be unsynthesizable by M1Score, confirming the need for RetroGNNScore. In line with early work in de novo design, we also observe that as the degrees of freedom increase, i.e. the size of the explorable chemical space gets larger from 105 to 464 fragments and then to graph edit search space, more negative $\text{AntibioticScore}(x)$ are seen, but these molecules tend to become increasingly chemically infeasible. To illustrate the differences in synthetic accessibility, chemical structures of top results are shown in Figure 4, and some proposed synthetic pathways by Molecule.one are shown in Figure 10.

[2] In the full version of this paper, we will provide a comparison against additional baseline models [3].

Table 1: Results on the task of finding synthesizable drug-like molecules with antibiotic properties, using different search strategies. In comparison to molecules found by screening the Zinc15 database, using RetroGNNScore results in easily synthesizable molecules that have higher predicted antibiotic potency. Cells show the values for the 1st (left) and 100th (right) top-scoring compounds.

(a) Results of 1000 searches in the 104 fragments search space using softmax action selection.

| Synthesizability score | Score (↓) | | AntibioticScore (↓) | | QED (↑) | | M1Score (↓) | | SAscore (↓) | |
|---|---|---|---|---|---|---|---|---|---|---|
| | Top-1st | Top-100th | Top-1st | Top-100th | Top-1st | Top-100th | Top-1st | Top-100th | Top-1st | Top-100th |
| RetroGNNScore | -4.581 | -3.333 | -4.581 | -3.364 | 0.725 | 0.781 | 3.671 | 4.882 | 2.676 | 3.205 |
| SAScore | -4.449 | -3.58 | -4.450 | -3.58 | 0.706 | 0.794 | 11 | 11 | 3.232 | 3.455 |

(b) Results of 1000 searches in the 485 fragments search space using softmax action selection.

| Synthesizability score | Score (↓) | | AntibioticScore (↓) | | QED (↑) | | M1Score (↓) | | SAscore (↓) | |
|---|---|---|---|---|---|---|---|---|---|---|
| | Top-1st | Top-100th | Top-1st | Top-100th | Top-1st | Top-100th | Top-1st | Top-100th | Top-1st | Top-100th |
| RetroGNNScore | -6.808 | -4.521 | -9.251 | -3.364 | 0.345 | 0.781 | 4.829 | 4.507 | 4.128 | 3.205 |
| SAScore | -6.994 | -3.714 | -10.993 | -3.761 | 0.490 | 0.734 | 11 | 11 | 4.187 | 3.594 |

(c) Results of 1000 searches in the graph edit search space using softmax action selection.

| Synthesizability score | Score (↓) | | AntibioticScore (↓) | | QED (↑) | | M1Score (↓) | | SAscore (↓) | |
|---|---|---|---|---|---|---|---|---|---|---|
| | Top-1st | Top-100th | Top-1st | Top-100th | Top-1st | Top-100th | Top-1st | Top-100th | Top-1st | Top-100th |
| RetroGNNScore | -4.641 | -3.658 | -4.641 | -3.857 | 0.814 | 0.696 | 5.921 | 4.260 | 3.382 | 3.697 |
| SAScore | -8.272 | -5.386 | -12.610 | -7.534 | 0.664 | 0.692 | 11 | 11 | 5.468 | 5.219 |

(d) Screening of 5.8M molecules in the Zinc15 database

| Synthesizability score | Score (↓) | | AntibioticScore (↓) | | QED (↑) | | M1Score (↓) | | SAscore (↓) | |
|---|---|---|---|---|---|---|---|---|---|---|
| | Top-1st | Top-100th | Top-1st | Top-100th | Top-1st | Top-100th | Top-1st | Top-100th | Top-1st | Top-100th |
| RetroGNNScore | -3.698 | -2.545 | -3.698 | -2.545 | 0.782 | 0.818 | 4.217 | 1.235 | 3.663 | 2.230 |
| SAScore | -3.617 | -2.597 | -3.698 | -2.597 | 0.782 | 0.700 | 4.217 | 3.690 | 3.663 | 3.266 |

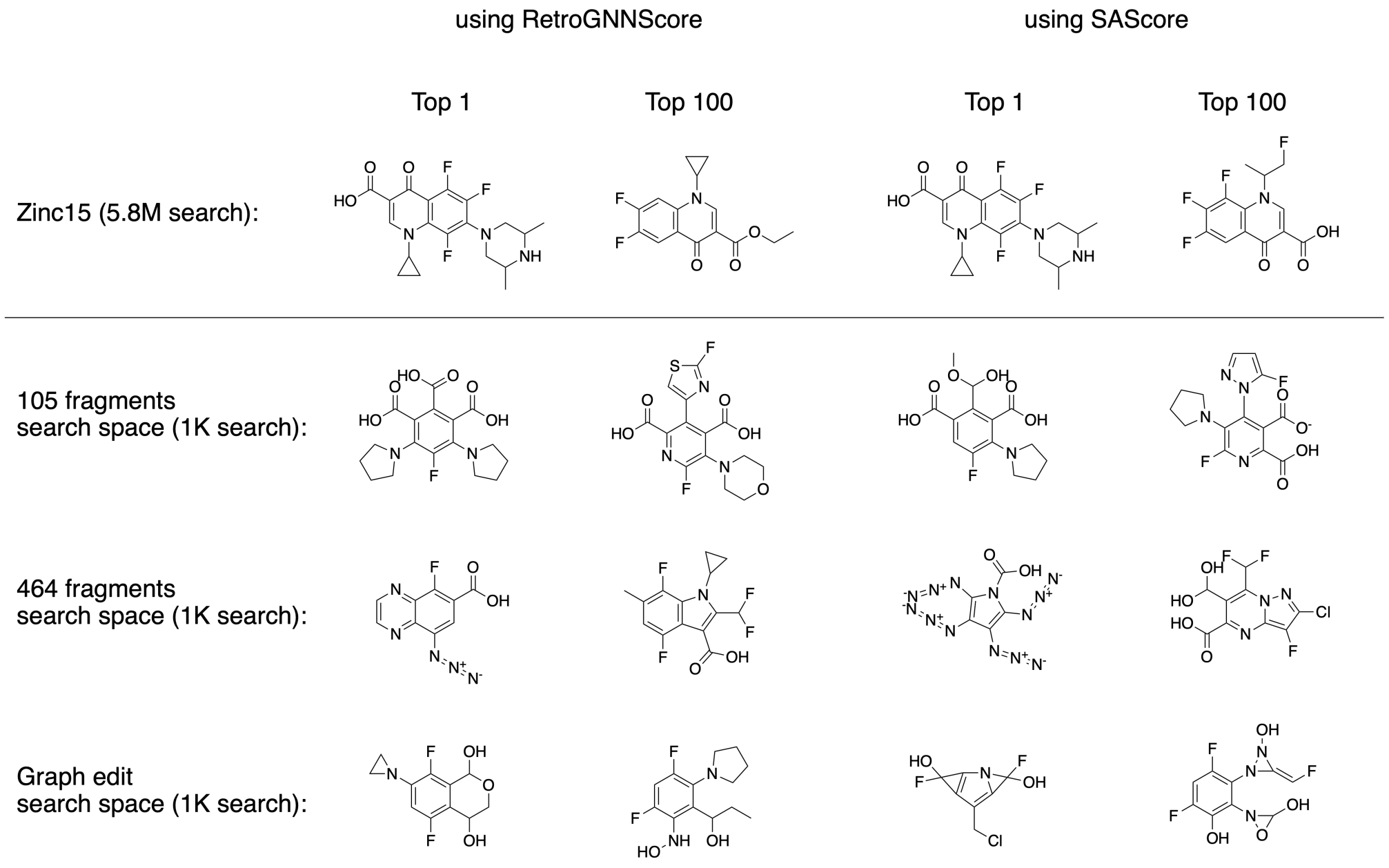

Figure 4: The use of RetroGNNScore results in predominantly synthesizable molecules whereas the use of SAScore does not. Also, the number of molecules exhibiting undesirable moieties is higher when using SAScore. Examples are shown via top-1st, and top-100th antibiotics candidates found in different search spaces.

## 4 Conclusion

Retrosynthesis planning software is the gold-standard to predict synthetic accessibility. However, this way of scoring is generally too slow to be used in de novo drug design. Instead, we propose to first train a deep neural network to approximate outputs of a retrosynthesis software, and then use the resulting RetroGNN to bias the search process.

We validated the developed synthetic accessibility score by searching for easy to synthesize molecules that can exhibit antibiotic properties. In particular, RetroGNN was able to filter out hard to synthesize molecules (with 0.99 AUC) while achieving a $10^5$ time speed-up over full retrosynthesis software.

Besides its use within de novo design, we note that our RetroGNNScore can be further interpreted and used as a value function or learned heuristic to guide computer-aided synthesis planning, as proposed in [19].

## References

[1] Jean-Louis Reymond, Lars Ruddigkeit, Lorenz Blum, and Ruud van Deursen. The enumeration of chemical space. *Wiley Interdisciplinary Reviews: Computational Molecular Science*, 2(5), 2012.

[2] Benjamin Sanchez-Lengeling and Alán Aspuru-Guzik. Inverse molecular design using machine learning: Generative models for matter engineering. *Science*, 361(6400), 2018.

[3] Nathan Brown, Marco Fiscato, Marwin H.S. Segler, and Alain C. Vaucher. GuacaMol: Benchmarking Models for de Novo Molecular Design. *Journal of Chemical Information and Modeling*, 59(3), 2019.

[4] Wenhao Gao and Connor W. Coley. The Synthesizability of Molecules Proposed by Generative Models. *Journal of Chemical Information and Modeling*, 2020.

[5] Krisztina Boda, Thomas Seidel, and Johann Gasteiger. Structure and reaction based evaluation of synthetic accessibility. *Journal of Computer-Aided Molecular Design*, 21(6), 2007.

[6] Connor W. Coley, Luke Rogers, W. Green, and K. Jensen. SCScore: Synthetic Complexity Learned from a Reaction Corpus. *Journal of Chemical Information and Modeling*, 58 2, 2018.

[7] Peter Ertl and Ansgar Schuffenhauer. Estimation of synthetic accessibility score of drug-like molecules based on molecular complexity and fragment contributions. *Journal of Cheminformatics*, 1(1), 2009.

[8] John Bradshaw, Brooks Paige, Matt J Kusner, Marwin Segler, and José Miguel Hernández-Lobato. Barking up the right tree: an approach to search over molecule synthesis DAGs. In *Advances in Neural Information Processing Systems 33*, 2020.

[9] John Bradshaw, Brooks Paige, Matt J Kusner, Marwin Segler, and José Miguel Hernández-Lobato. A Model to Search for Synthesizable Molecules. In *Advances in Neural Information Processing Systems 32*, 2019.

[10] Julien Horwood and Emmanuel Noutahi. Molecular Design in Synthetically Accessible Chemical Space via Deep Reinforcement Learning, 2020. arXiv:2004.14308.

[11] Sai Krishna Gottipati, Boris Sattarov, Sufeng Niu, Yashaswi Pathak, Haoran Wei, Shengchao Liu, Karam M. J. Thomas, Simon Blackburn, Connor W. Coley, Jian Tang, Sarath Chandar, and Yoshua Bengio. Learning to navigate the synthetically accessible chemical space using reinforcement learning. In *International Conference on Machine Learning*, 2020.

[12] Tomasz Klucznik, Barbara Mikulak-Klucznik, Michael P. McCormack, Heather Lima, Sara Szymkuć, Manishabrata Bhowmick, Karol Molga, Yubai Zhou, Lindsey Rickershauser, Ewa P. Gajewska, Alexei Toutchkine, Piotr Dittwald, Michał P. Startek, Gregory J. Kirkovits, Rafał Roszak, Ariel Adamski, Bianka Sieredzińska, Milan Mrksich, Sarah L. J. Trice, and Bartosz A. Grzybowski. Efficient Syntheses of Diverse, Medicinally Relevant Targets Planned by Computer and Executed in the Laboratory. *Chem*, 4(3), 2018.

[13] Marwin H. S. Segler, Thierry Kogej, Christian Tyrchan, and Mark P. Waller. Generating Focused Molecule Libraries for Drug Discovery with Recurrent Neural Networks. *ACS Central Science*, 4(1), 2018.

[14] Samuel Genheden, Amol Thakkar, Veronika Chadimova, Jean-Louis Reymond, Ola Engkvist, and Esben Jannik Bjerrum. AiZynthFinder: A Fast Robust and Flexible Open-Source Software for Retrosynthetic Planning. *Journal of Cheminformatics*, 12(70), 2020.

[15] Marwin HS Segler. World programs for model-based learning and planning in compositional state and action spaces, 2019. arXiv:1912.13007.

[16] Molecule.one. Available at `https://molecule.one`.

[17] Jonathan M. Stokes, Kevin Yang, Kyle Swanson, Wengong Jin, Andres Cubillos-Ruiz, Nina M. Donghia, Craig R. MacNair, Shawn French, Lindsey A. Carfrae, Zohar Bloom-Ackermann, Victoria M. Tran, Anush Chiappino-Pepe, Ahmed H. Badran, Ian W. Andrews, Emma J. Chory, George M. Church, Eric D. Brown, Tommi S. Jaakkola, Regina Barzilay, and James J. Collins. A Deep Learning Approach to Antibiotic Discovery. *Cell*, 180(4), 2020.

[18] Binghong Chen, Chengtao Li, Hanjun Dai, and Le Song. Retro*: Learning Retrosynthetic Planning with Neural Guided A* Search. In *International Conference on Machine Learning*, 2020.

[19] Marwin H. S. Segler, Mike Preuss, and Mark P. Waller. Planning chemical syntheses with deep neural networks and symbolic AI. *Nature*, 555(7698), 2018.

[20] Justin Gilmer, Samuel S Schoenholz, Patrick F Riley, Oriol Vinyals, and George E Dahl. Neural Message Passing for Quantum Chemistry. In *Proceedings of Machine Learning Research 70*, 2017.

[21] Kevin Yang, Kyle Swanson, Wengong Jin, Connor Coley, Philipp Eiden, Hua Gao, Angel Guzman-Perez, Timothy Hopper, Brian Kelley, Miriam Mathea, Andrew Palmer, Volker Settels, Tommi Jaakkola, Klavs Jensen, and Regina Barzilay. Analyzing Learned Molecular Representations for Property Prediction. *Journal of Chemical Information and Modeling*, 59(8), 2019.

[22] Qi Liu, Miltiadis Allamanis, Marc Brockschmidt, and Alexander Gaunt. Constrained Graph Variational Autoencoders for Molecule Design. In *Advances in Neural Information Processing Systems 31*, 2018.

[23] Wengong Jin, Regina Barzilay, and Tommi Jaakkola. Hierarchical Generation of Molecular Graphs using Structural Motifs, 2020. arXiv:2002.03230.

[24] Niclas St\aahl, Göran Falkman, Alexander Karlsson, Gunnar Mathiason, and Jonas Boström. Deep Reinforcement Learning for Multiparameter Optimization in de novo Drug Design. *Journal of Chemical Information and Modeling*, 59(7), 2019.

[25] Nicolò Cesa-Bianchi, Claudio Gentile, Gabor Lugosi, and Gergely Neu. Boltzmann Exploration Done Right. In *Advances in Neural Information Processing Systems 30*, 2017.

[26] G. Richard Bickerton, Gaia V. Paolini, Jérémy Besnard, Sorel Muresan, and Andrew L. Hopkins. Quantifying the chemical beauty of drugs. *Nature Chemistry*, 4(2), 2012.

[27] Teague Sterling and John J. Irwin. ZINC 15 – Ligand Discovery for Everyone. *Journal of Chemical Information and Modeling*, 55(11), 2015.

[28] Petar Veličković, Guillem Cucurull, Arantxa Casanova, Adriana Romero, Pietro Liò, and Yoshua Bengio. Graph Attention Networks. In *International Conference on Learning Representations*, 2018.

[29] Daniel Mark Lowe. *Extraction of Chemical Structures and Reactions from the Literature*. PhD Thesis, University of Cambridge, 2012.

[30] Helen M. Berman, John Westbrook, Zukang Feng, Gary Gilliland, T. N. Bhat, Helge Weissig, Ilya N. Shindyalov, and Philip E. Bourne. The Protein Data Bank. *Nucleic Acids Research*, 28(1), 2000.

## Acknowledgements

This work was supported by IVADO. CL and MK want to thank WL Hamilton, K Lapchevskyi, M Jain, and B Rosseau (Mila) for helpful discussions. MS wants to thank JL Reymond (University of Bern) for enlightening discussions.

## Appendix

### 4.1 Molecule.one synthetic accessibility score

On a high level, the version of M1Score used in this paper is calculated using A* search algorithm that finds a path from the target molecule to purchasable starting materials [18, 19]. The search is guided using (1) a binary classifier that predicts the probability of a reaction succeeding, and (2) a heuristic that approximates the price of an intermediate compound. The classifier is modeled using Graph Attention Neural Network [28] and trained on chemical reactions from the US patent office [29]. In each search, maximally 10,000 nodes (a node corresponds to an intermediate molecule) are visited. The final score combines classifier predictions with compound prices fetched from the eMolecules database.

In contrast, SAScore is based on the estimation of the complexity of the individual fragments and the overall molecules. Compared to M1Score, the lack of consideration for retrosynthesis in SAScore biases it to underestimate the availability of commercial molecules (e.g. in FDA-approved drugs) and overestimate the synthetic accessibility of new molecules, especially in a block-based environment (Figure 5).

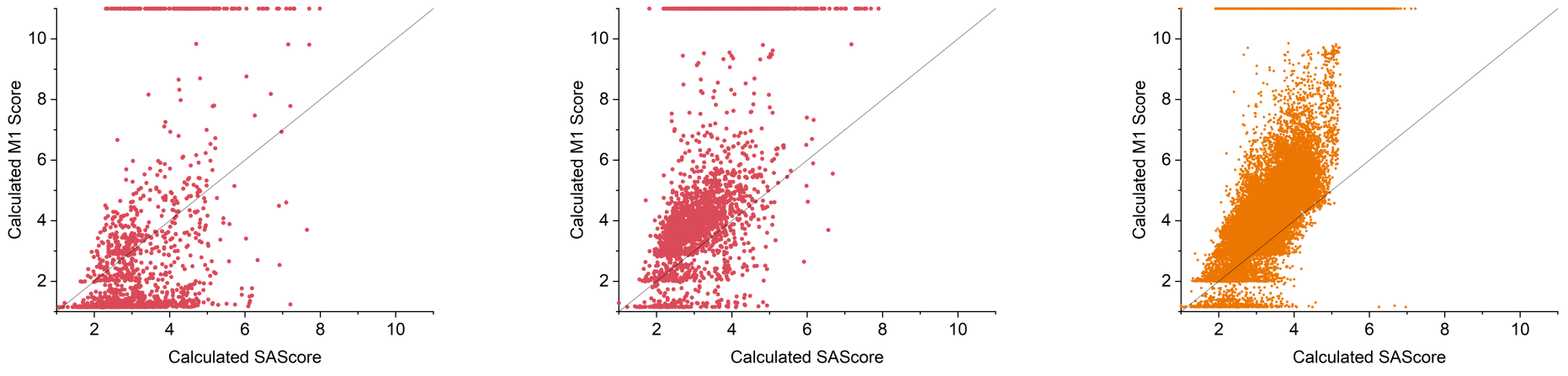

(a) FDA-approved drugs (b) Investigational compounds (c) Random molecules from 105 fragment search space

Figure 5: Comparison of SAScore with M1Score to show M1Score estimates synthesizability better. In FDA-approved drugs, M1Score considers the commercial availability of the compound, hence SAScore underestimates the synthetic accessibility; in investigational drugs from Zinc15 database and randomly generated molecules, SAScore overestimates synthetic accessibility by lack of retrosynthesis consideration.

### 4.2 Experimental details on RetroGNNScore

RetroGNNScore is trained on implementations of both [20] and [21], separately for each search space with 50,000 M1 scores calculated from randomly generated molecules; we also tried training from the combination of three search spaces, which gives slightly worse $R^2$ (<0.05 difference) than when trained separately. Bayesian optimization was used in each case to find the optimal hidden size, depth, and dropout; generally, a hidden size of 1000, depth of 6, and dropout of 0.05 lead to good performance. The two implementations produce similar results in terms of $R^2$, and for consistency, all results in this paper come from implementations of [21]. Regression and binary classification were carried out with 5-fold cross-validation. An example ROC curve can be found in Figure 7.

### 4.3 Experimental details on de novo design of antibiotics

In this paper, we consider three search spaces, which we depict in Figure 6. In the first two search spaces, we consider the construction of a molecule to be from connecting or disconnecting building blocks. The available fragments are chosen based on the frequency of appearance in the ligand subset of PDB database[30], and we picked the top 105 and 464 fragments respectively. In the third search space, we experiment with the editing of graphs, where we allow simple, chemically-sensible, and versatile actions such as adding/editing/deleting an atom, editing/deleting a bond, and fusing an aliphatic or aromatic ring. We cap the maximum number of non-hydrogen atoms at 50.

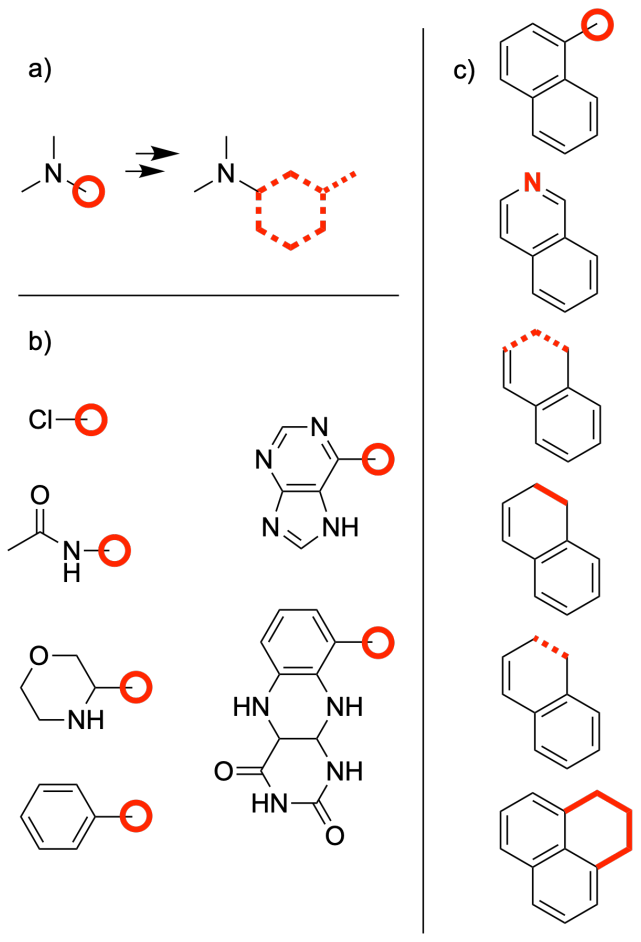

Figure 6: Illustration of atom-based (**a**), fragment-based (**b**), and graph-based (**c**) search spaces, where we explore (**b**) and (**c**) in the paper.

Molecular design in this paper is based on soft-max action selection (Equation 6), where $P_i(t)$ is the probability of choosing action $i$ on trial $t$, and $Q_i(t)$ is the score of the succeeding state. Here, the softmax effect is achieved by the Boltzmann distribution, and the positive temperature $\tau$ controls the amount of exploration in the vicinity of good-performing actions (exploration versus exploitation). A high temperature ($\tau \to \infty$) leads to an equal probability of choice of actions, and a low temperature ($\tau \to 0$) biases the action towards greedy exploitation. 0.15 was determined to be the optimal temperature for all three search spaces. Note that each search is initialized with 20 random actions.

$$P_i(t) = \frac{\exp\left(Q_i(t)/\tau\right)}{\sum_{j=1}^{n} \exp\left(Q_j(t)/\tau\right)} \tag{6}$$

### 4.4 Additional experimental results

We here include a series of additional results that support our paper. This includes a comparison of speed between RetroGNN and Molecule.one (Table 2), results for RetroGNN binary classification (Figure 7), correlation plots between RetroGNNScore and M1Score for known drug-like molecules (Figure 8), comparisons of scores obtained via screening known drug-like molecules, molecules generated by random search, and molecules generated by softmax action selection (Figure 9), and synthetic pathways proposed by Molecule.one for top-scoring molecules found by using RetroGNNScore in our search (Figure 10).

| Method | Time needed/molecule |
|---|---|
| RetroGNNScore | 0.8 ms |
| M1Score | 202 s |

Table 2: Time comparison of calculating RetroGNNScore (average of 100,000, on a GPU) and M1Score (average of 120, by Molecule.one (non-exploratory) public API on November 2020, on a single CPU core).

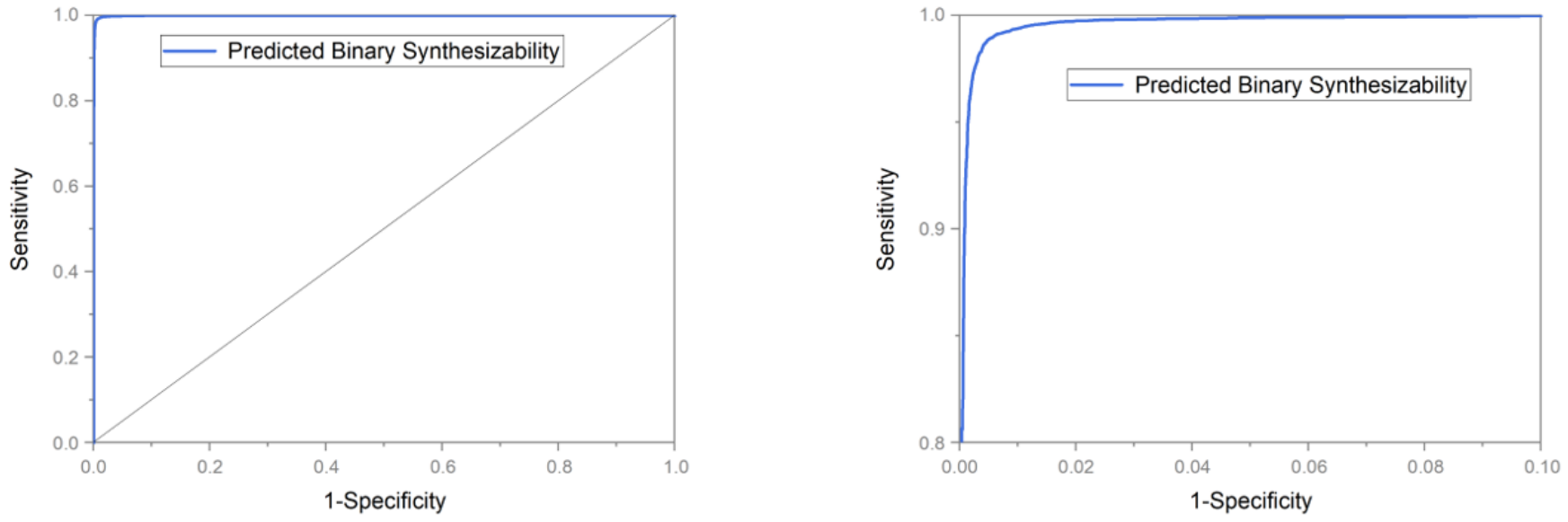

Figure 7: RetroGNNScore classifies well whether a compound has any synthetic routes by M1Score. Example receiver operating characteristic curve (left) and its zoom in (right) on 50,000 random molecules generated in search space of 105 fragments

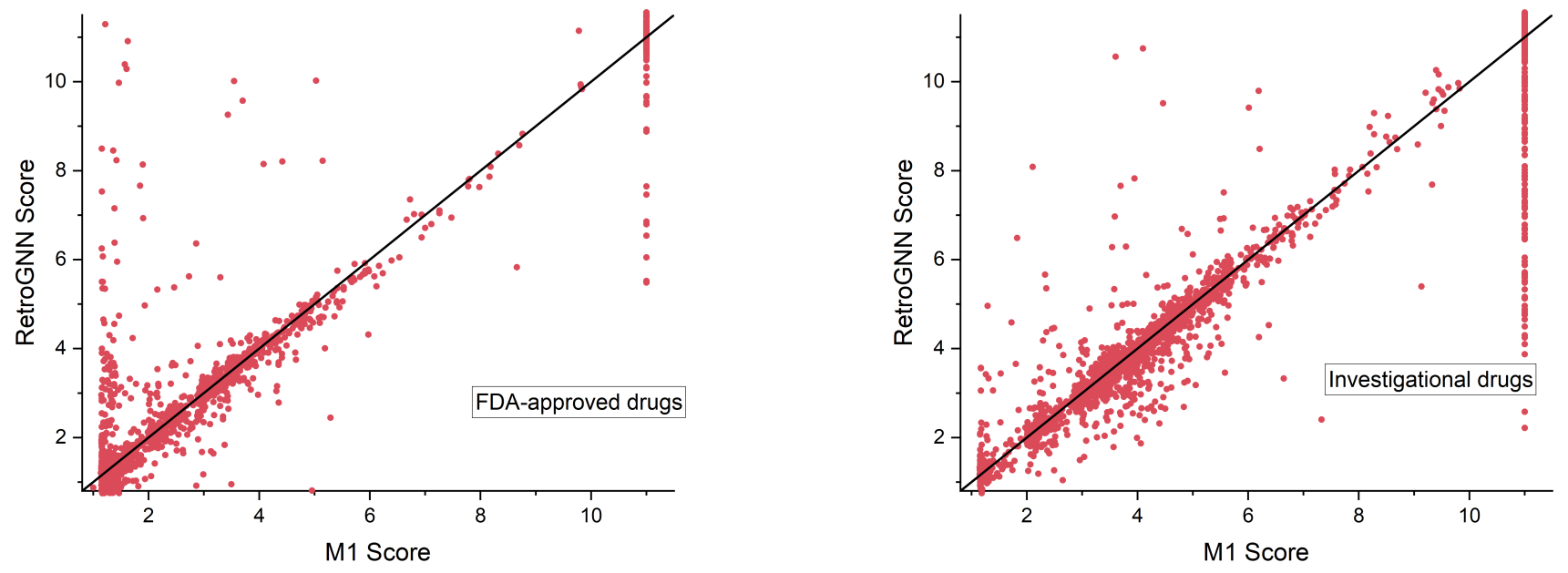

(a) FDA-approved drugs

(b) Investigational drugs

Figure 8: RetroGNNScore approximates M1Score well on FDA-approved and investigational drugs, albeit with slightly lower $R^2 = 0.910, 0.941$, respectively. We suspect the lower correlation originates from the absence of similar molecules in the training dataset and other factors such as more commercial availability despite being highly complex molecules as they are used as drugs in real life.

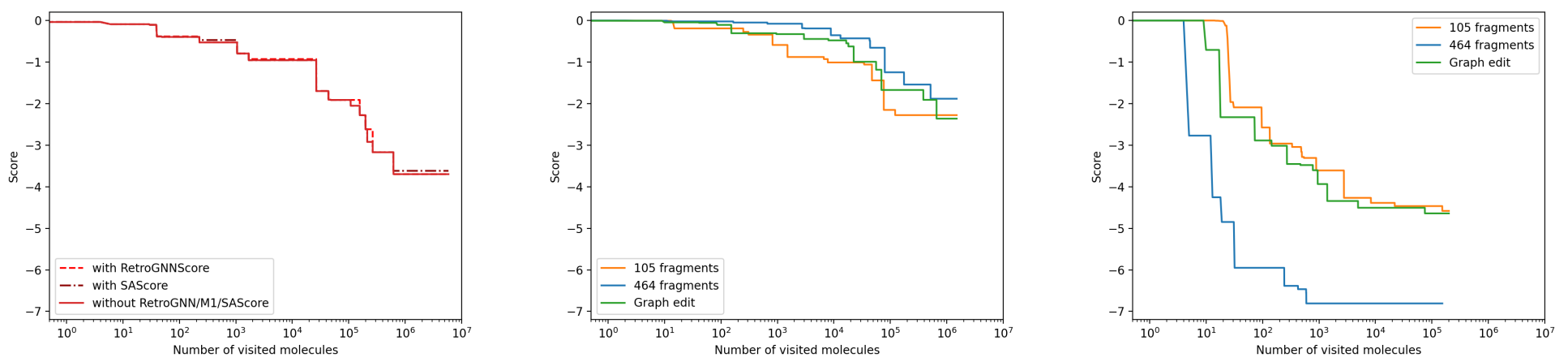

(a) Enumerating Zinc15 database

(b) Enumerating random molecules

(c) Enumerating molecules from softmax action selection

Figure 9: Softmax action selection (**c**) outperforms enumeration of known drug-like molecules (Zinc15 database, **a**) and randomly-generated molecules (**b**). Here the lowest score thus far is plotted against the number of visited molecules; M1Score is approximated by RetroGNNScore; the objective is optimizing Equation 1.

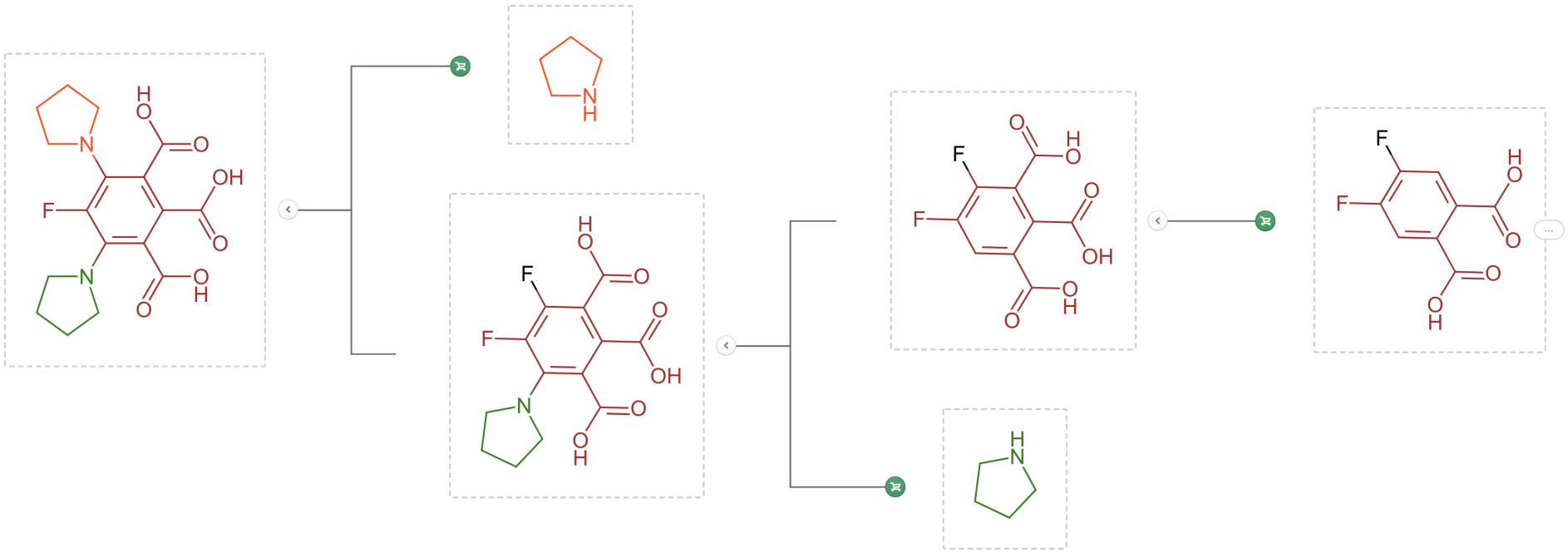

(a) 105 fragments search space

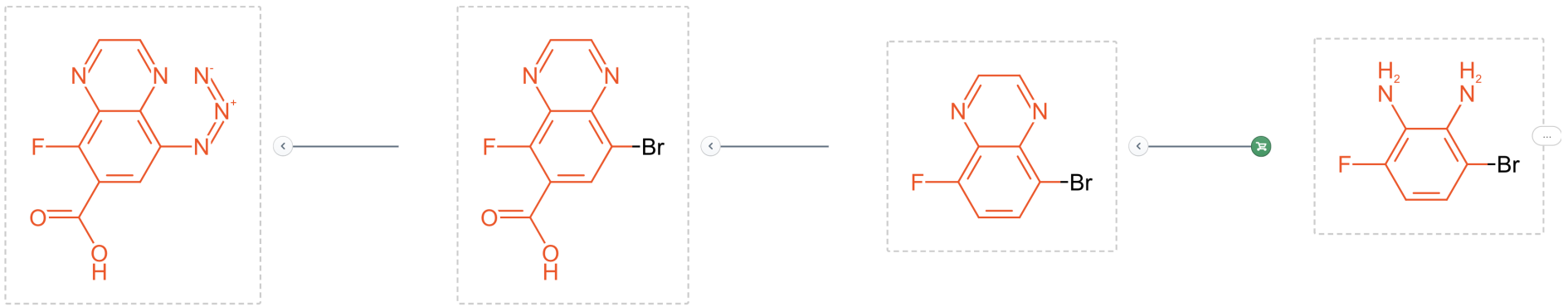

(b) 464 fragments search space

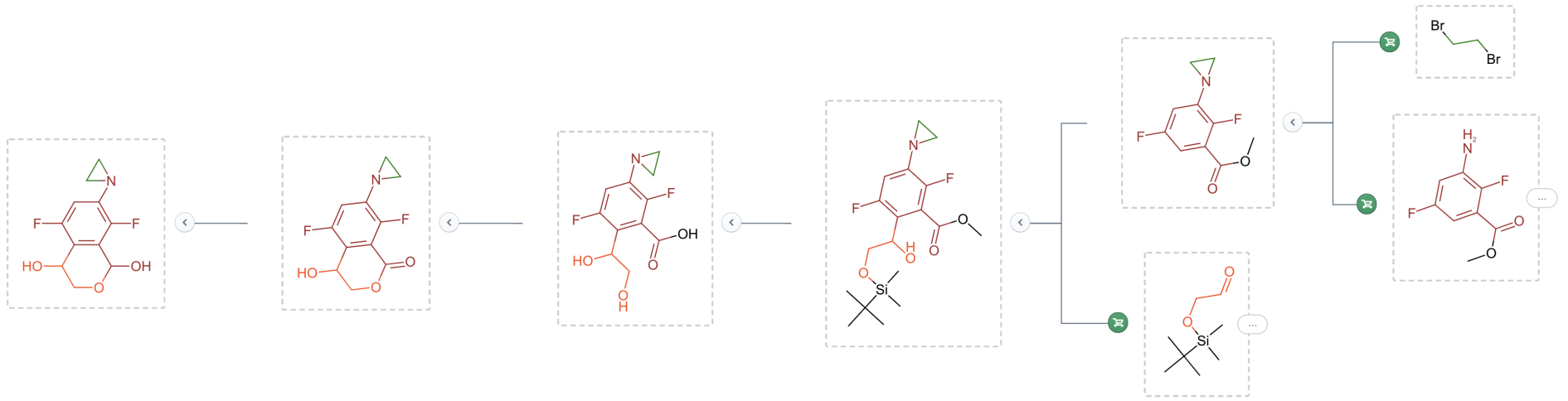

(c) Graph edit search space

Figure 10: Synthetic pathways proposed by Molecule.one for the top-1st antibiotics candidate found in each search space using softmax action selection.